\documentclass[10pt,twocolumn,letterpaper]{article}

\usepackage{wacv}              

\definecolor{wacvblue}{rgb}{0.21,0.49,0.74}
\usepackage[pagebackref,breaklinks,colorlinks,allcolors=wacvblue]{hyperref}

\usepackage[table]{xcolor}
\usepackage{pifont}
\usepackage[accsupp]{axessibility}
\newcommand{\cmark}{\ding{51}}%
\newcommand{\xmark}{\ding{55}}%
\usepackage{multirow}

\newcommand{\update}[1]{\textcolor{black}{#1}}

\def\wacvPaperID{834} 
\def\confName{WACV}
\def\confYear{2027}

\title{Fieldable Low-Light Multi-Task Perception via Foundation-Model-Assisted Domain Adaptation}

\author{{Beomseok Kang$^{\dag}$}\thanks{Equal contribution. BK primarily contributed to FAMDA method development and standard-benchmark evaluation; NM primarily contributed to the fieldable low-light system, dataset, and field evaluation. Most of BK’s contributions were made during an internship at SRI International.}~, Niluthpol Chowdhury Mithun$^{\ddag}$\footnotemark[1]~,\\Mikhail Sizintsev$^{\ddag}$,~Han-Pang Chiu$^{\ddag}$,~Supun Samarasekera$^{\ddag}$ 
\and
{\normalsize $^{\dag}$Seoul National University, Seoul, South Korea} \and {\normalsize $^{\ddag}$SRI International, Princeton, NJ, USA}
\and
\hspace{1.1cm} \texttt{\small $^{\dag}$beomseok@snu.ac.kr} \hspace{1.8cm} \and \texttt{\small $^{\ddag}$firstname.lastname@sri.com}
}

\begin{document}
\maketitle
\begin{abstract}
Fieldable low-light perception systems require both semantic and geometric scene understanding under severe domain shift and embedded compute constraints. We present FAMDA, a foundation-model-assisted self-training framework for unsupervised domain adaptation (UDA) in multi-task dense prediction. The framework integrates segmentation and depth foundation models into a self-training paradigm to generate high-quality pseudo-labels for the target domain, distilling their generalization capabilities into a single lightweight student network. To connect the method to deployment, we instantiate the adapted student on a compact handheld low-light prototype with DomiNite imaging and Jetson Orin Nano embedded compute. We evaluate on standard synthetic-to-real multi-task UDA benchmarks for comparison with prior work and on our field-collected low-light dataset to validate daytime-to-low-light adaptation in the target application setting. Across these evaluations, the proposed approach achieves strong segmentation and depth accuracy with substantially lower memory use and inference latency than direct foundation-model inference, demonstrating a practical path toward embedded multi-task perception for low-light field systems. 
Code: \url{https://github.com/beomseokg/FAMDA}.
\end{abstract}
    
\section{Introduction}
\label{sec:intro}

Fieldable low-light perception systems require reliable scene understanding under adverse illumination, severe domain shift, and embedded compute constraints. Applications such as outdoor navigation, surveillance, situational awareness, and AR-assisted operator support benefit from both semantic and geometric understanding of the environment. Multi-task dense prediction addresses this need by jointly solving pixel-level tasks such as semantic segmentation and depth estimation~\cite{vandenhende2021multi, choi2024multi}. Semantic segmentation provides categorical scene understanding, while depth estimation recovers geometric structure. Sharing these tasks within a single network can improve representation learning~\cite{ghiasi2021multi} and reduce computation through shared feature extraction~\cite{sun2020adashare}, making multi-task learning attractive for resource-constrained field systems.

\begin{figure}[t]
\centering
\vspace{0.2cm}
\includegraphics[width=0.99\linewidth]{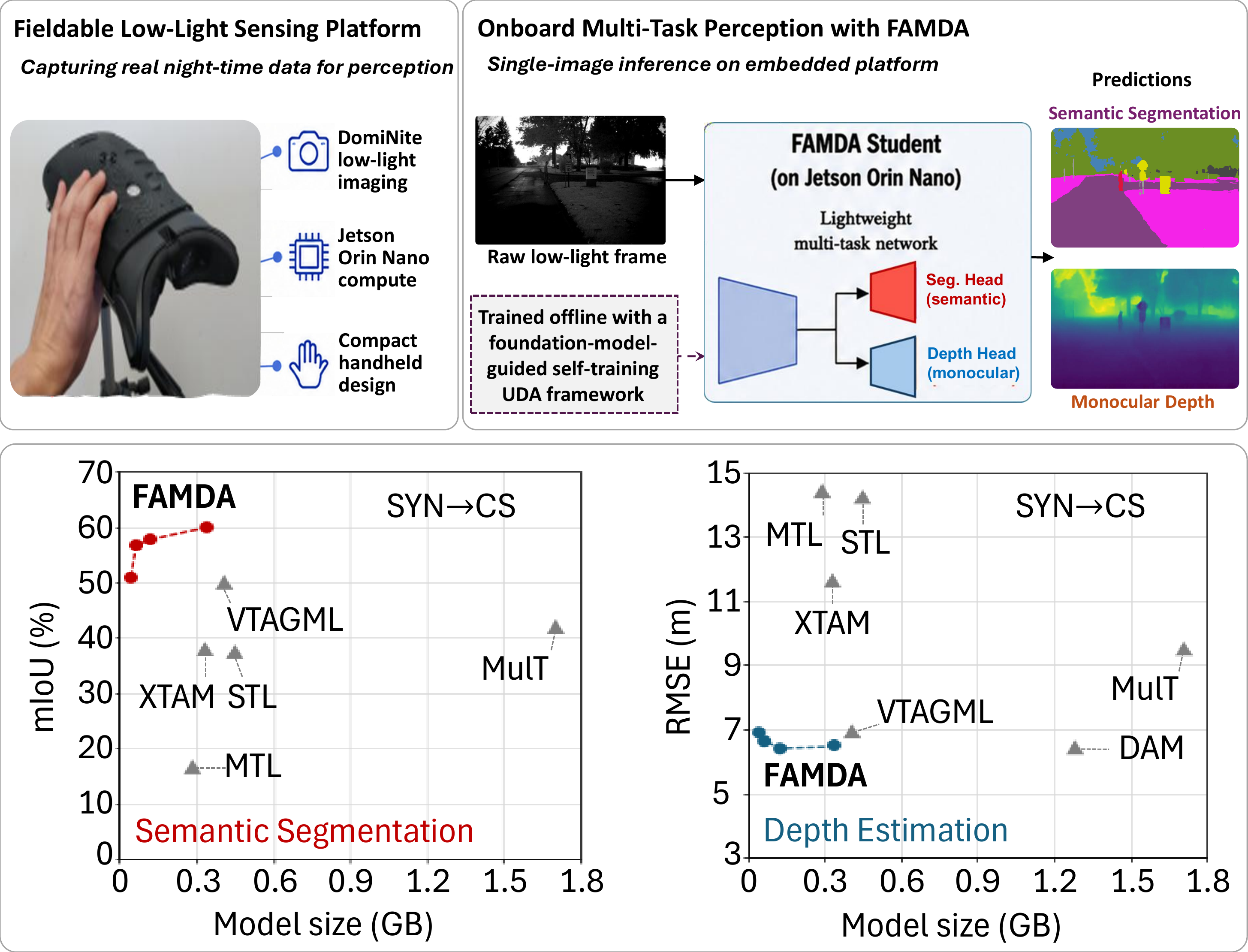}
\vspace{-0.2cm}
\caption{
\textbf{Fieldable low-light multi-task perception.}
Top: our compact low-light sensing platform and onboard FAMDA pipeline, where a lightweight student model predicts semantic segmentation and monocular depth from a single low-light frame.
Bottom: accuracy-efficiency tradeoff on SYNTHIA$\rightarrow$Cityscapes, comparing model size against segmentation mIoU and depth RMSE with prior work. FAMDA provides strong performance with compact multi-task models.
}
\label{fig:teaser}
\vspace{-0.4cm}
\end{figure}

However, supervised multi-task learning is difficult to support for new environments~\cite{lu2020multi}. Dense semantic labels require costly manual annotation, while depth supervision often depends on stereo, multi-view reconstruction, or specialized sensors. These challenges are amplified in low-light settings, where sensor noise, illumination, contrast, and glare differ substantially from common daytime training data. Unsupervised domain adaptation provides a practical way to adapt from labeled source data to unlabeled target imagery~\cite{kundu2019adapt, lopes2023cross, zhao2024unsupervised, karim2023c}, but multi-task UDA remains less explored than single-task adaptation. The few existing methods mainly rely on adversarial learning~\cite{bhattacharjee2022mult, bhattacharjee2023vision}, whereas recent advances in UDA, especially for semantic segmentation, are largely driven by self-training~\cite{ hoyer2023mic, kang2026duda}. Extending self-training to multi-task prediction is nontrivial: pseudo-labels must be reliable for multiple tasks, depth prediction introduces scale and geometry constraints, and lightweight teacher models often degrade under large domain shifts. 

Vision foundation models (VFMs) can bridge this gap between adversarial multi-task UDA and self-training methods that have so far succeeded mainly in single-task segmentation.
Segmentation and depth foundation models have shown strong generalization across diverse visual domains~\cite{ravi2024sam, zhao2023fast, yin2023metric3d, ke2025marigold, yang2024depth2}, but their memory and latency costs make direct deployment difficult on compact embedded platforms. This motivates using VFMs as training-time teachers rather than deployment-time models.

In this paper, we present \textbf{FAMDA}, \textbf{F}oundation-model-\textbf{A}ssisted \textbf{M}ulti-task \textbf{D}omain \textbf{A}daptation, a self-training framework for efficient multi-task dense prediction. During adaptation, it integrates segmentation and depth foundation models into the pseudo-labeling process, distilling their generalization capabilities into a single lightweight student network. At inference, only the student is used, predicting semantic segmentation and monocular depth from one image. 
To connect the method to a fieldable low-light application, we instantiate FAMDA on a compact handheld low-light prototype with DomiNite low-light imaging~\cite{dominite} and Jetson Orin Nano compute (Fig.~\ref{fig:teaser}). We collect a nighttime outdoor dataset using the same platform and evaluate segmentation accuracy, depth accuracy, and onboard runtime under the conditions the system is meant to operate in. We further evaluate FAMDA on standard synthetic-to-real multi-task UDA benchmarks, including SYNTHIA$\rightarrow$Cityscapes and Virtual KITTI2$\rightarrow$Cityscapes, for direct comparison with prior works. This evaluation reflects the current benchmark landscape: established multi-task UDA benchmarks provide both segmentation and depth evaluation, while low-light domain adaptation benchmarks primarily focus on semantic segmentation. 

The key contributions are summarized as follows:

\begin{itemize}


\item We propose FAMDA, a framework that integrates vision foundation models into self-training-based UDA for multi-task learning, enabling effective knowledge distillation from large-scale pre-trained models.


\item We show that FAMDA provides strong accuracy-efficiency tradeoffs on standard synthetic-to-real multi-task UDA benchmarks, with substantially lower memory use and latency than direct foundation-model inference.

\item We validate on a field-collected low-light dataset using a compact handheld prototype, demonstrating daytime-to-low-light adaptation for embedded multi-task perception.

\end{itemize}
\section{Related Works}

\label{sec:related}

\textbf{Multi-task perception }jointly estimates multiple pixel-level outputs, such as semantic segmentation and depth estimation, from a shared visual representation~\cite{vandenhende2021multi, choi2024multi}. This formulation is useful for fieldable perception systems because semantic segmentation provides categorical scene understanding, while depth estimation provides geometric layout. Sharing computation across tasks can also reduce memory and inference cost compared to running separate task models~\cite{sun2020adashare}, which is important for embedded systems operating under latency, power, and form-factor constraints.


Prior work has largely focused on architecture and optimization: encoder- and decoder-side designs for sharing and separating task features~\cite{lopes2023cross, liu2019end, bruggemann2020automated, yang2024multi, xu2018pad, sinodinos2025cross, zhang2019pattern, bhattacharjee2022mult, li2022learning, zamir2020robust}, task-balancing schemes~\cite{yu2020gradient, kendall2018multi, chen2018gradnorm}, and depth as an auxiliary cue for domain-adaptive segmentation~\cite{wang2021domain, sick2024unsupervised}. We instead focus on the training paradigm for adapting multi-task dense prediction to unlabeled target domains while preserving deployment efficiency.

\begin{figure*}[t]
\centering
\includegraphics[width=0.88\textwidth]{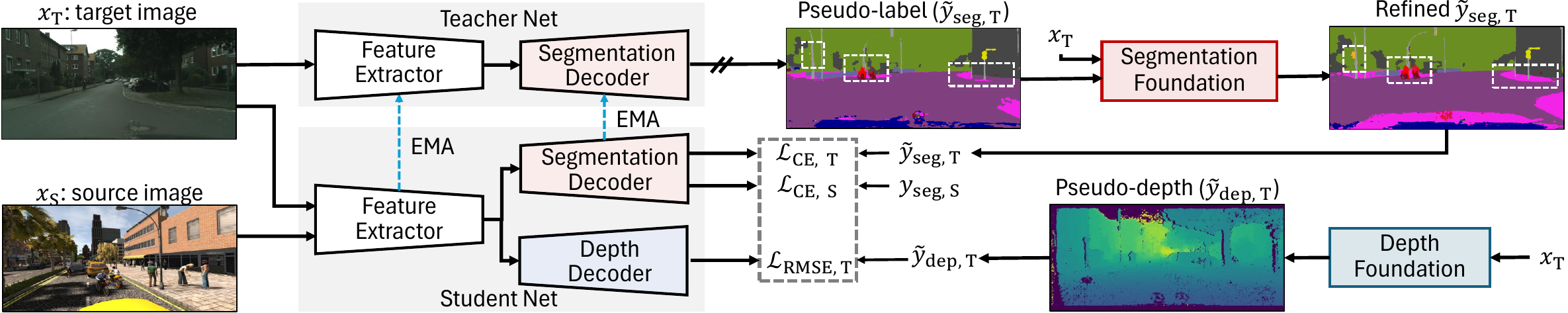}
\vspace{-0.2cm}
\caption{
\textbf{Overview of FAMDA.}
FAMDA uses a self-training framework for multi-task UDA. A teacher network generates target-domain semantic pseudo-labels, which are refined using masks from a segmentation foundation model. A depth foundation model provides pseudo-depth supervision for target images. \update{Both foundation models remain frozen.} The resulting segmentation and depth supervision are distilled to a single lightweight student network\update{, which alone is used at deployment.}
}
\label{fig:overview}
\vspace{-0.2cm}
\end{figure*}

\noindent \textbf{Unsupervised Domain Adaptation} for multi-task dense prediction has been explored in a few prior works (\textit{e.g.}, UM-adapt~\cite{kundu2019adapt}, XTAM~\cite{lopes2023cross}, and VTAGML~\cite{bhattacharjee2023vision}), which primarily rely on adversarial alignment and contribute decoder architectures such as cross-task distillation, attention, and interaction. 
Meanwhile, state-of-the-art UDA for semantic segmentation have shifted toward self-training~\cite{hoyer2022daformer,hoyer2022hrda,hoyer2023mic}, where a teacher network generates target-domain pseudo-labels to supervise a student network, but these advances have largely remained single-task. 

Several works incorporate SAM into adaptation, including SAM4UDASS~\cite{yan2023sam4udass}, illumination adaptation with SAM~\cite{mu2025illumination}, and SAM-guided 3D segmentation~\cite{cao2024mopa,yang2025sam}. These methods demonstrate the value of foundation-model guidance, but primarily focus on semantic segmentation. FAMDA instead integrates both segmentation and depth foundation models into self-training for joint semantic and geometric adaptation, complementing prior architectural designs rather than replacing them, with the option to add further dense tasks such as surface normals.


\noindent \textbf{Low-light and nighttime scene understanding } has been widely studied for adverse-condition perception. Datasets such as Dark Zurich~\cite{sakaridis2020map}, NightCity~\cite{tan2021night}, and ACDC~\cite{sakaridis2021acdc} support evaluation under challenging illumination, weather, and nighttime conditions. Existing low-light UDA studies primarily focus on semantic segmentation, largely because public low-light benchmarks usually provide semantic labels but not paired depth annotations. This leaves a gap for low-light multi-task dense perception, where both semantic categories and scene geometry are needed for navigation, surveillance, and AR-style operator support. 

\noindent \textbf{Vision Foundation Models } have shown strong zero-shot generalization across diverse domains. Segment Anything~\cite{kirillov2023segment} provides strong class-agnostic segmentation priors, while depth foundation models such as Depth Anything~\cite{yang2024depth,yang2024depth2}, and Marigold~\cite{ke2025marigold} provide robust monocular depth estimates. However, direct inference with these models is costly, particularly when multiple dense prediction tasks must run simultaneously on embedded platforms. FAMDA therefore uses segmentation and depth foundation models as training-time teachers, distilling their guidance into a single lightweight student network for deployment.


\section{Method}
\label{sec:method}

\subsection{Overview}
\label{subsec:overview}

We propose \textbf{FAMDA}, a foundation-model-assisted self-training framework for UDA in multi-task dense prediction. Given labeled source images $\{x_S, y^{seg}_S\}$ and unlabeled target images $\{x_T\}$, our goal is to train a single model $f_{\theta}$ that predicts both semantic segmentation and monocular depth in the target domain. The model consists of a shared encoder and task-specific decoder heads:
\begin{equation}
\vspace{-0.2cm}
\small
f_{\theta}(x) = \left(f^{seg}_{\theta}(x), f^{dep}_{\theta}(x)\right),
\end{equation}
where $f^{seg}_{\theta}(x) \in \mathbb{R}^{H \times W \times C}$ and $f^{dep}_{\theta}(x) \in \mathbb{R}^{H \times W}$ denote semantic segmentation and depth predictions, respectively.

FAMDA builds on EMA-based self-training. A teacher network generates target-domain semantic pseudo-labels, while the student is optimized using source labels and target pseudo-labels. The teacher is updated as an exponential moving average of the student: $\theta_T \leftarrow \alpha \theta_T + (1-\alpha)\theta_S$,
where $\theta_T$ and $\theta_S$ are teacher and student parameters, and $\alpha$ is the EMA momentum. Since depth pseudo-labels are provided directly by a depth foundation model, the EMA teacher is used only for the shared feature extractor and semantic segmentation decoder. 

\update{Overall, UDA defines the labeled-source/unlabeled-target setting, EMA self-training provides the semantic pseudo-labeling scaffold, and foundation models provide training-time guidance that is discarded at inference. Thus, SAM and DAM are not additional deployed models or components of the EMA teacher; they are external training-time pseudo-supervision sources, and only the compact student is retained at inference.
}

\subsection{Foundation-Model-Assisted Pseudo-Labeling}
\label{subsec:fm-assisted-labeling}

\noindent\textbf{Semantic segmentation.}
For target-domain semantic supervision, the teacher network first predicts semantic pseudo-labels on unlabeled target images. Since Segment Anything (SAM)~\cite{kirillov2023segment} produces class-agnostic object masks rather than semantic labels, we use SAM to refine the teacher predictions. Following prior SAM-guided UDA methods~\cite{yan2023sam4udass,yang2025sam}, SAM masks are generated offline and cached. During training, each SAM mask is assigned the majority semantic class predicted by the teacher within that mask. This produces refined target pseudo-labels $\tilde{y}^{seg}_T$ with improved spatial consistency and object boundaries.

\noindent\textbf{Depth estimation.}
For target-domain depth supervision, we use Depth Anything (DAM)~\cite{yang2024depth} to generate pseudo-depth maps $\tilde{y}^{dep}_T$ for target images. Unlike SAM, DAM directly outputs dense monocular depth predictions and therefore does not require refinement through the task teacher. This provides a simple way to extend self-training-based UDA, originally developed mainly for semantic segmentation, to joint semantic and geometric adaptation. Importantly, source-domain depth labels are not required, making the framework applicable even when only source semantic labels and unlabeled target images are available.

\subsection{Training Objective}

FAMDA incorporates three loss functions: cross-entropy (CE) loss for semantic segmentation on source (\(\mathcal{L_\text{CE, S}}\)) and target (\(\mathcal{L_\text{CE, T}}\)) domains, and root mean square error (RMSE) loss for depth estimation on the target domain (\(\mathcal{L_\text{RMSE, T}}\)). For semantic segmentation, the CE losses penalize classification errors on both labeled source images and pseudo-labeled target images:
\vspace{-0.1cm}
\begin{equation}
\small
\mathcal{L}_{\text{CE, S}}=-\sum_{j=1}^{H \times W}\sum_{c=1}^{C}{y_\text{S}^{(j, c)} \cdot \log{f_{\theta}}(x_{\text{S}})^{(j, c)}}
\label{eq:ce_source}
\end{equation}
\vspace{-0.2cm}
\begin{equation}
\small
\mathcal{L}_{\text{CE, T}}=-\sum_{j=1}^{H \times W}\sum_{c=1}^{C}{\tilde{y}_\text{T}^{(j, c)} \cdot \log{f_{\theta}(x_{\text{T}})^{(j, c)}}},
\label{eq:ce_target}
\end{equation}
\noindent where $f_{\theta}$ denotes the student model with a \textit{segmentation} decoder, \(y_\text{S}\) is true labels in the source, and \(\tilde{y}_\text{T}\) represents the pseudo-labels generated by the teacher model on the target.


We use a scale- and shift-invariant depth loss, as DAM provides relative depth pseudo-labels.
~\cite{ranftl2020towards}:
\vspace{-0.1cm}
\begin{equation}
\small
    \mathcal{L_\text{RMSE, T}}=\text{RMSE}(\text{SSI}(f_{\theta}(x_{\text{T}})), \text{SSI}(\tilde{y}_{\text{T}})),
\label{eq:ssi_rmse}
\end{equation}
\begin{equation}  
\small
\mathrm{SSI}(y) =
\frac{y-\mathrm{med}(y)}
{\mathrm{mean}\left(|y-\mathrm{med}(y)|\right)} .
\end{equation}
\noindent where \(f_{\theta}\) is the student with a \textit{depth} decoder, and \(\tilde{y}_\text{T}\) is pseudo-depth from DAM. This normalization stabilizes training from relative depth pseudo-labels and is used only during training. We define $\mathcal{L}_{seg}=\mathcal{L}_{CE,S}+\mathcal{L}_{CE,T}$ and $\mathcal{L}_{dep}=\mathcal{L}_{RMSE,T}$. The final objective is
\begin{equation}
\vspace{-0.1cm}
\mathcal{L}_{total} = \mathcal{L}_{seg} + \beta \mathcal{L}_{dep},
\end{equation}
where $\beta$ balances semantic and depth supervision. Scale-and-shift invariant normalization is used only for training with relative pseudo-depth; the benchmark RMSE protocol and metric-scale calibration are described in Sec.~\ref{sec:exp_setup}.

\begin{figure}[t]
\centering
\includegraphics[width=0.99\linewidth]{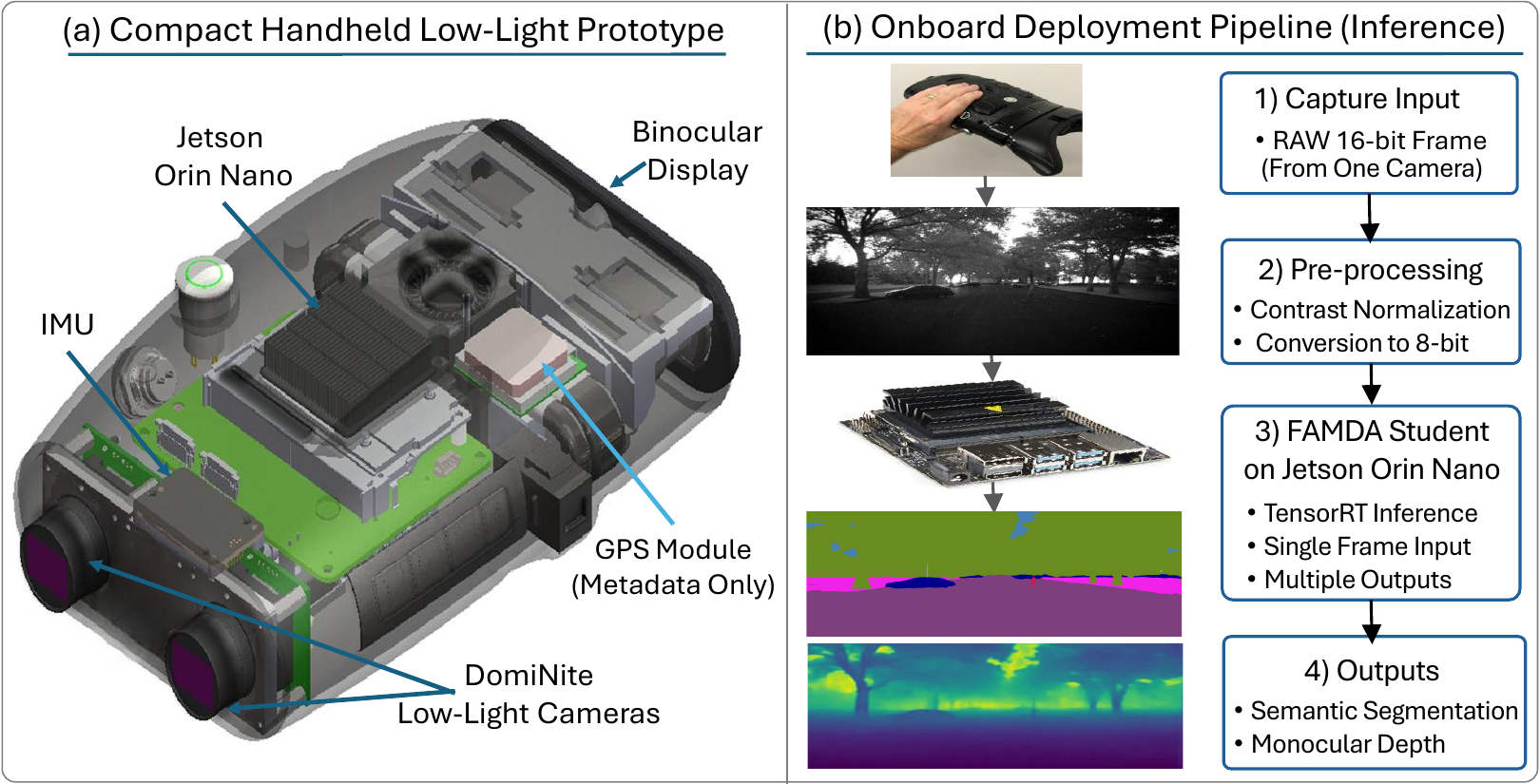}
\vspace{-0.2cm}
\caption{
\textbf{Fieldable low-light multi-task perception pipeline.}
(a) Compact handheld prototype with DomiNite low-light imaging and embedded Jetson compute for onboard multi-task perception.
(b) Onboard deployment pipeline: raw 16-bit low-light frames are preprocessed before being passed to the deployed FAMDA student model. At inference time, the student predicts semantic segmentation and monocular depth from a single frame. 
}
\label{fig:deployment_pipeline}
\vspace{-0.3cm}
\end{figure}

\subsection{Inference Model}

After adaptation, the segmentation foundation model, depth foundation model, and EMA teacher are discarded. Only the compact student network is used at inference time. Given a single input image, the student predicts semantic segmentation and monocular depth in one forward pass. This design allows FAMDA to use large foundation models as training-time teachers while avoiding their memory and latency costs during deployment.
\section{Fieldable Low-Light System}
\label{sec:deployment}



\subsection{Prototype and Onboard Inference Pipeline}

Fig.~\ref{fig:deployment_pipeline}(a) shows our handheld low-light prototype. To connect FAMDA to a fieldable low-light application, we instantiate the adapted student model on a compact handheld system with DomiNite low-light imaging \cite{dominite} and embedded Jetson compute. \update{The DomiNite camera is a high-sensitivity visible-band sensor that produces usable imagery at night without active or infrared illumination, so the device is carried by a dismounted operator to see and navigate in unlit environments.} The prototype is designed for outdoor nighttime perception under practical size, weight, and power constraints. In addition to the imaging and compute modules, the system integrates power regulation and a low-power display for operator-facing visualization. 

The prototype currently uses a calibrated two-camera configuration for low-light data collection. Initially designed to support visual-inertial sensing, this setup also enables stereo-derived depth reference generation for evaluation, as described later in Sec.~\ref{sec:lowlight_depth_ref}. However, FAMDA itself is not a stereo method. The deployed student network performs single-image inference: given one low-light frame, it jointly predicts semantic segmentation and monocular depth in a single forward pass. 
This design also motivates future single-camera versions of the platform, which could reduce cost, size, power consumption, synchronization requirements, and calibration complexity. 


The deployed system produces semantic segmentation and depth outputs that can be visualized as operator-facing overlays (\textit{e.g.}, categorical cues from segmentation, geometric layout and relative range information from depth). These outputs support low-light situational awareness and navigation assistance under embedded compute constraints. The system is intended as a human-in-the-loop perception aid and does not currently make autonomous decisions.


\subsection{Onboard Pipeline}
\label{sec:onboard_deployment}

\noindent\textbf{Fast low-light preprocessing.}
The DomiNite imaging system provides high-bit-depth low-light frames, whereas the adapted FAMDA student uses an ImageNet-pretrained backbone and expects standard image statistics. Each 16-bit frame is converted to an 8-bit input using clipped linear contrast normalization: intensities are clipped to limits $(l,u)$, linearly rescaled to $[0,255]$, cast to 8-bit, resized, and normalized. This deterministic linear operation is selected over more expensive enhancement methods to keep embedded preprocessing latency low. \update{The limits $(l,u)$ are estimated once from unlabeled target-domain training frames available under UDA and then fixed for target training, evaluation, and deployment; no target labels, stereo references, or test-set ground truth are used. For a new camera, exposure setting, or deployment domain, $(l,u)$ should be re-estimated from a short unlabeled calibration stream, which we treat as a preprocessing dependency rather than an optimized low-light enhancement step.} Applying the same processing to target training, manually annotated evaluation, and deployment frames keeps image statistics consistent across adaptation and deployment.

\noindent\update{\textbf{RAW/high-bit-depth tradeoff.}
This preprocessing is a practical deployment choice rather than an optimal low-light image representation. Converting high-bit-depth sensor data to 8-bit enables compatibility with pretrained backbones and keeps the preprocessing deterministic and low latency, but it can discard weak signal and sensor statistics that may be useful under extreme darkness. Prior RAW/noise-aware low-light recognition work, including \emph{Instance Segmentation in the Dark}~\cite{chen2023instance}, shows that low-light sensor noise and high-frequency feature disturbances can degrade dense recognition, and that RAW/high-bit-depth inputs can provide useful information beyond 8-bit images. Therefore, the low-light accuracy reported here should be interpreted as the performance of FAMDA under a lightweight 8-bit deployment front end. RAW-aware encoders, learned ISP/front-end processing, and feature-denoising are complementary future directions.}

\noindent\textbf{Model export and onboard inference.}
The deployed student model consists of a shared encoder with task-specific decoder heads for semantic segmentation and depth estimation, allowing both dense outputs to be produced in one forward pass from a single input frame. The trained PyTorch student model is exported to ONNX at a fixed input resolution. The ONNX model is then transferred to the Jetson device and converted to a TensorRT engine on the target platform. Building the TensorRT engine on the deployment device allows the runtime to optimize the model for the target GPU, precision mode, and available memory. Fig.~\ref{fig:deployment_pipeline}(b) summarizes the onboard deployment pipeline from 16-bit image capture to output visualization.

\begin{table}[t]
\centering
\renewcommand{\arraystretch}{0.94}
\caption{Prototype and deployment configuration.}
\label{tab:prototype_config}
\vspace{-0.2cm}
\resizebox{\linewidth}{!}{
\begin{tabular}{l l}
\toprule
Component & Configuration \\
\midrule
Imaging & DomiNite low-light camera system ($65^\circ$ HFOV) \\
Camera configuration & Calibrated two-camera setup for data collection/evaluation \\
Network input & Single low-light frame \\
Predicted outputs & Semantic segmentation and monocular depth \\
Embedded compute & Jetson Orin Nano \\
Deployment format & PyTorch $\rightarrow$ ONNX $\rightarrow$ TensorRT \\
Input format & 16-bit low-light frame \\
Preprocessing & Contrast normalization to 8-bit, resize, normalize \\
Foundation models & Used only for training or pseudo-label generation \\
Stereo processing & Used only for depth-reference generation/evaluation \\
IMU and GPS & Metadata logging only; not used for perception \\
Display & Low-power visualization display \\
Power envelope & $\sim$15 W \\
Form factor & $8.7 \times 5.4 \times 2.2$ inches \\
Weight & $\sim$3 lbs \\
\bottomrule
\end{tabular}}
\vspace{-0.35cm}
\end{table}

\section{Experiments and Field Evaluation}
\label{sec:experiments}

We evaluate FAMDA in two complementary settings: daytime-to-low-light adaptation on our field-collected dataset (Sec.~\ref{sec:lowlight_eval}), covering semantic segmentation, monocular depth prediction, and embedded runtime; and standard synthetic-to-real multi-task UDA benchmarks (Sec.~\ref{sec:standard_benchmarks}).


\subsection{Evaluation and Implementation Details}
\label{sec:exp_setup}

\noindent\textbf{Metrics.}
We report mIoU for semantic segmentation and RMSE for depth. On synthetic-to-real benchmarks, RMSE is computed on depth maps after median scaling to the ground-truth depth following prior multi-task UDA and monocular depth evaluation protocols~\cite{lopes2023cross,godard2019digging}, excluding pixels outside $[10^{-3},80]$ m. On the field-collected low-light evaluation, it is computed over valid pixels from the stereo-derived depth reference described in Sec.~\ref{sec:lowlight_depth_ref}. We provide detailed RMSE formulas in the supplementary material.

\noindent\textbf{Metric scale calibration.}
Median scaling is useful for benchmark comparability, but it requires ground-truth depth unavailable at deployment. For optional deployment-oriented metric depth, we estimate a global scale and shift between FAMDA relative depth and a metric depth foundation model using $|\mathcal{X}_{cal}|{=}20$ unlabeled target-domain calibration images, disjoint from all evaluation images. We use the DAM-V2-Large outdoor fine-tuned model~\cite{yang2024depth2} only for this calibration; it is not used to train FAMDA, generate target-domain pseudo-depth during adaptation, or construct evaluation references. 
\update{Thus, the procedure is target-ground-truth-free, using no semantic labels, stereo references, or measured metric depth, but not foundation-model-free because it relies on a metric depth VFM pseudo-reference and a small unlabeled calibration set.}


\noindent\textbf{Training implementation.}
We use Segment Anything~\cite{kirillov2023segment} with a ViT-H encoder for semantic pseudo-label refinement and Depth Anything~\cite{yang2024depth} with a ViT-L encoder for target-domain pseudo-depth generation. We use 128 points per side, prediction IoU threshold 0.86, and stability score threshold 0.92,  for SAM-based mask generation. Our training setup follows DAFormer~\cite{hoyer2022daformer}, including ImageNet pretraining, feature-distance loss, rare-class sampling, and standard augmentations. We use source-target image mixing for semantic segmentation, but not for depth supervision because pasted objects do not share the same viewpoint or scene geometry as the target image. For the depth branch, we therefore use geometry-consistent augmentations such as color jittering, cropping, and flipping. We extend the DAFormer dataloaders to support Virtual KITTI2 and our field-collected low-light data.

\noindent\textbf{Backbones and model variants.}
FAMDA is not tied to a particular network architecture. For the main accuracy-efficiency analysis and low-light deployment study, we use SegFormer backbones (MiT-B0, B1, B2, B5) with DAFormer decoder heads in parallel for segmentation and depth. For comparison with prior multi-task UDA work, we also evaluate DeepLab-V2 with a ResNet-101 backbone, modifying its decoder to more closely match the DAFormer structure. We use DAFormer rather than heavier extensions such as HRDA~\cite{hoyer2022hrda} and MIC~\cite{hoyer2023mic} to balance UDA accuracy with low-latency embedded deployment.

\subsection{Field-Collected Low-Light Evaluation}
\label{sec:lowlight_eval}

\subsubsection{Data Collection and Semantic Annotation}
\label{sec:lowlight_data}

To evaluate FAMDA in the target low-light application setting, we collect nighttime outdoor imagery using our low-light camera system~\cite{dominite}. The source domain is Cityscapes daytime imagery, and the target domain is our unlabeled low-light data. This creates a daytime-to-low-light adaptation setting that reflects the deployment scenario of fieldable perception under adverse illumination.

The data are collected in semi-urban outdoor environments (\textit{e.g.}, campus roads, sidewalks, parking areas, dirt roads, buildings, vegetation, and open low-light regions) in two phases: June sequences captured between 8:30--10:30 PM during handheld walking and cart-based collection, and September sequences captured between 7:30--9:00 PM from a vehicle. The sequences include substantial variation in illumination, viewpoint, motion, and sensor settings. In particular, we collect scenes with most streetlights turned off as well as scenes with active streetlights, vehicle sequences with and without headlight glare, and data captured under different exposure/gain and lens/focus settings. These variations reflect challenging imaging conditions encountered by fieldable low-light perception systems.

\begin{table}[t]
\centering
\renewcommand{\arraystretch}{1.05}
\setlength{\tabcolsep}{4pt}
\fontsize{8.4pt}{8.4pt}\selectfont
\caption{
\textbf{Field-collected low-light dataset evaluation.}
FAMDA provides both semantic segmentation and monocular depth from one model. Metrics are mIoU (\%, $\uparrow$) over 13 annotated classes and RMSE (m, $\downarrow$) over valid stereo-reference pixels. Model size and A5000 latency are reported for model-level comparison; embedded TensorRT runtime is reported in Tab.~\ref{tab:embedded_runtime}.
}
\vspace{-0.2cm}
\label{tab:lowlight}
\resizebox{\linewidth}{!}{
\begin{tabular}{l|c|cc|cc}
\toprule
\textbf{Method} & \textbf{Outputs} 
& \cellcolor{pink!25}\textbf{mIoU} 
& \cellcolor{cyan!10}\textbf{RMSE}
& \textbf{Model Size} & \textbf{A5000 Lat.} \\
& & (\%) & (m) & (MB) & (ms) \\
\midrule
SSAM-H        & Seg.        & 43.93 & --   & 3351 & 34117 \\
DAM-L         & Depth       & --    & 5.47 & 1279 & 60.8  \\
DaF-Seg-B5    & Seg.        & 52.92 & --   & 325  & 35.5  \\
DaF-Dep-B5    & Depth       & --    & 5.55 & 325  & 36.7  \\
DaF-Seg/Dep-B5$^\dagger$ & Seg.+Depth & 52.92 & 5.55 & 650 & 72.2 \\
\midrule
Ours-B0       & Seg.+Depth & 43.02 & 5.54 & 40   & 20.6 \\
Ours-B1       & Seg.+Depth & 49.08 & 5.54 & 79   & 22.5 \\
Ours-B2       & Seg.+Depth & 54.72 & 5.53 & 121  & 28.9 \\
Ours-B5       & Seg.+Depth & 55.32 & 5.53 & 339  & 42.8 \\
\bottomrule
\multicolumn{6}{l}{$^\dagger$ Estimated by combining separate single-task models under sequential execution.}
\end{tabular}}
\vspace{-0.4cm}
\end{table}

For semantic segmentation evaluation, we manually annotate 51 frames using Labelbox. The labeled frames are selected across different locations, lighting conditions, and collection modes to reduce temporal redundancy. We use Cityscapes-compatible semantic classes to maintain consistency with the daytime source-domain training setup. In the low-light evaluation, mIoU is computed over 13 annotated classes: road, sidewalk, building, wall, fence, pole, traffic light, traffic sign, vegetation, terrain, sky, person, and car. From the remaining unlabeled low-light frames, 1,000 images are used as the target-domain training set for UDA.

\subsubsection{Stereo-Derived Depth Reference}
\label{sec:lowlight_depth_ref}

Although FAMDA predicts monocular depth from a single image, the calibrated two-camera DomiNite setup provides a practical way to obtain depth references for low-light evaluation. We calibrate the camera pair using a standard checkerboard calibration procedure~\cite{zhang2000calibration}, rectify the stereo images, and compute disparity at the native image resolution using FoundationStereo~\cite{wen2025stereo}. The resulting disparity maps are converted to depth using the calibrated camera intrinsics and baseline.

The stereo-derived depth maps are used only as evaluation references and are not used as target-domain ground-truth supervision during UDA training. Since stereo matching can be unreliable in saturated regions, very dark regions, low-texture areas, reflective surfaces, sky, and far-range regions, invalid disparity pixels and pixels outside the valid depth range are masked before computing RMSE.

\begin{figure*}[t]
\centering
\vspace{0.1cm}
\includegraphics[width=0.88\linewidth, height=3.6cm]{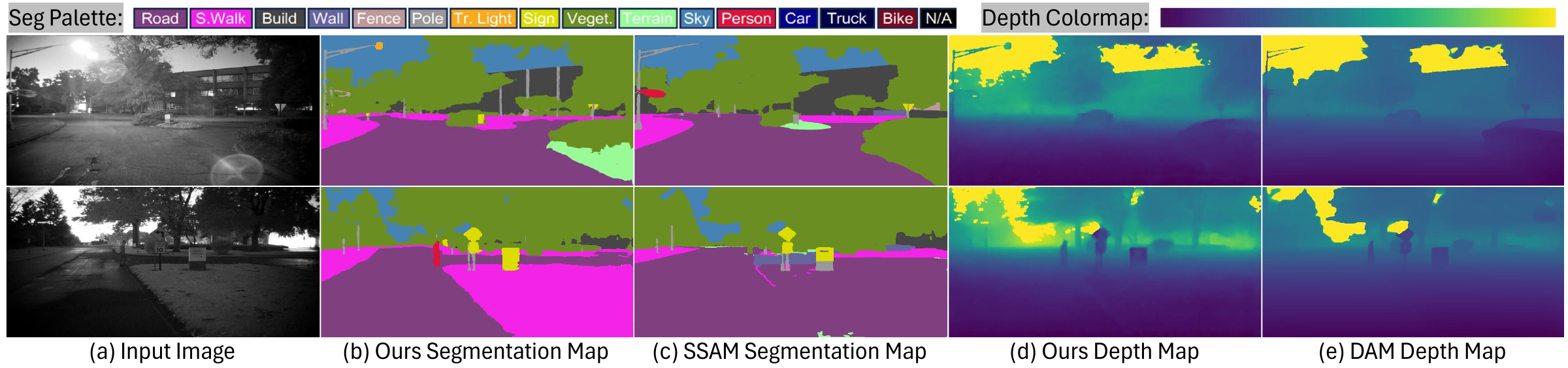}
\vspace{-0.3cm}
\caption{
\textbf{Qualitative results on the field-collected low-light dataset.}
We compare our model with single-task foundation-model baselines: SSAM for semantic segmentation and DAM for depth estimation. Leveraging its joint outputs, FAMDA assigns the maximum depth to pixels segmented as sky. FAMDA produces both outputs from a single MiT-B5 model.
}
\vspace{-0.3cm}
\label{fig:lowlight_qualitative}
\end{figure*}

\subsubsection{Baselines and Metrics}
\label{sec:lowlight_baselines}


We compare FAMDA with both foundation-model baselines and single-task UDA baselines. For foundation models, we use Depth Anything with a ViT-L backbone (DAM-L) for depth estimation and Semantic Segment Anything (SSAM-H)~\cite{chen2023semantic,kirillov2023segment} for semantic segmentation. For single-task UDA, we train separate DAFormer-based models for semantic segmentation and depth estimation, denoted DaF-Seg and DaF-Dep. DaF-Dep is a lightweight single-task depth student trained with DAM-generated target pseudo-depth supervision, rather than direct DAM inference. For FAMDA, we evaluate MiT-B0, MiT-B1, MiT-B2, and MiT-B5 variants to study the accuracy-efficiency tradeoff.

We report mIoU for semantic segmentation over the 13 annotated low-light classes and RMSE for depth estimation over valid stereo-reference pixels. Tab.~\ref{tab:lowlight} reports model size and single-image inference latency on an A5000 GPU to compare FAMDA against large foundation-model and single-task baselines under the same software setting. Embedded TensorRT runtime on the Jetson Orin Nano is reported separately in Tab.~\ref{tab:embedded_runtime}.

\subsubsection{Low-Light Results}
\label{sec:lowlight_results}

Tab.~\ref{tab:lowlight} shows that FAMDA provides a strong segmentation-depth operating point on the field-collected low-light dataset. For semantic segmentation, Ours-B5 achieves 55.32 mIoU and Ours-B2 achieves 54.72 mIoU, outperforming the single-task DaF-Seg-B5 and SSAM-H segmentation foundation-model baselines. This indicates that explicit daytime-to-low-light adaptation is more effective than direct zero-shot semantic foundation-model inference in this setting. \update{Per-class, grouped, and bootstrap reliability analyses are reported in Suppl. (Tabs.~S5, S6, S7, S8).}

\begin{table}[t]
\centering
\renewcommand{\arraystretch}{1.0}
\setlength{\tabcolsep}{3.5pt}
\fontsize{8.0pt}{8.2pt}\selectfont
\caption{
\textbf{Embedded TensorRT runtime and low-light metrics on Jetson Orin Nano.}
\update{Runtime is measured on a Jetson Orin Nano 8GB using TensorRT FP16 at input resolution $526{\times}840$. 
Latency is the standardized core runtime used for comparison across backbones, excluding file I/O, output saving, overlay blending, and display update. A representative Ours-B2 component-level timing breakdown, including raw-frame preprocessing and display-resolution colorization, is provided in the supplementary (Tab.~S12). Peak memory reports RAM footprint above the idle baseline. Power denotes total board draw (VDD\_IN) during sustained inference. mIoU and RMSE use the same field-collected low-light labels and stereo references as Tab.~\ref{tab:lowlight}.}
}
\vspace{-0.2cm}
\label{tab:embedded_runtime}
\resizebox{\linewidth}{!}{
\begin{tabular}{l|ccc|cc|cc}
\toprule
\textbf{Model}
  & \textbf{TRT Size} & \textbf{Lat.} & \textbf{FPS}
  & \textbf{Peak Mem.} & \textbf{Power}
  & \cellcolor{pink!25}\textbf{mIoU} & \cellcolor{cyan!10}\textbf{RMSE} \\
  & (MB) & (ms) & (Hz) & (MB) & (W) & (\%) & (m) \\
\midrule
Ours-B0 & 23.4  & 123 & 8.1 & 1346 & 7.28 & 42.90 & 5.54 \\
Ours-B1 & 42.5  & 144 & 6.9 & 1361 & 7.54 & 49.23 & 5.53 \\
Ours-B2 & 66.0  & 158 & 6.3 & 1406 & 7.75 & 55.01 & 5.53 \\
Ours-B5 & 187.0 & 212 & 4.7 & 1629 & 8.80 & 55.96 & 5.52 \\
\bottomrule
\end{tabular}}
\vspace{-0.4cm}
\end{table}

For depth estimation, FAMDA variants achieve RMSE values around 5.53 m, comparable to DAM-L and the single-task DaF-Dep-B5 model. DAM-L obtains slightly lower RMSE, but it provides only depth and requires a substantially larger model. In contrast, FAMDA produces both semantic segmentation and depth from a single compact model. The MiT-B2 variant provides a favorable model-level operating point, achieving 54.72 mIoU and 5.53 m RMSE with 121 MB model size and 28.9 ms A5000 latency, while MiT-B5 provides the highest accuracy when additional compute is available. \update{Additional low-light depth metrics and valid-pixel coverage are reported in the supplementary (Tab.~S3). The valid stereo-reference coverage is about 1\%, so the low-light depth results should be interpreted as preserving useful geometric prediction in a compact joint model rather than as a large target-domain depth-accuracy improvement over DAM-L.}

The optional metric-scale calibration described in Sec.~\ref{sec:exp_setup} reduces Ours-B5 RMSE from 5.53 m to 4.50 m on the low-light evaluation. This calibration uses only unlabeled calibration images; no manually annotated semantic labels or stereo-derived depth references are used.
This result suggests that a simple global scale-and-shift calibration can improve metric-depth consistency in the target low-light domain without target ground-truth depth.

Fig.~\ref{fig:lowlight_qualitative} provides qualitative results with SSAM-H and DAM-L. FAMDA produces segmentation maps that better preserve low-light scene structure and minority classes such as traffic lights and pedestrians, while its depth predictions are visually aligned with DAM-L. 
\update{Failure cases remain in severe glare, very dark or low-texture regions, sky, and far-range areas, where stereo-derived references are sparse or unreliable and monocular predictions can also degrade.}

\begin{table}[t]
    \centering
    \renewcommand{\arraystretch}{1.05}
    \setlength{\tabcolsep}{6pt}
    \fontsize{8.4pt}{8.4pt}\selectfont
    \caption{\textbf{Comparison on standard synthetic-to-real multi-task UDA benchmarks.}
    The \textbf{best} and \underline{second-best} UDA results are highlighted. Metrics are mIoU (\%, $\uparrow$) and RMSE (m, $\downarrow$). Baseline results for STL, MTL, and prior UDA methods are cited from XTAM~\cite{lopes2023cross} and VTAGML~\cite{bhattacharjee2023vision}.}
    \vspace{-0.2cm}
    \label{tab:main_comparison}
    \resizebox{\linewidth}{!}{
    \begin{tabular}{c|c|c|cc|cc}
        \toprule
        & \multirow{2}{*}{\textbf{Method}} & \multirow{2}{*}{\textbf{Backbone}} 
        & \multicolumn{2}{c|}{\textbf{SYN$\rightarrow$CS} (16 class)} 
        & \multicolumn{2}{c}{\textbf{VK2$\rightarrow$CS} (8 class)} \\
        \cmidrule(r){4-7}
        & & & \cellcolor{pink!25}mIoU & \cellcolor{cyan!10}RMSE 
              & \cellcolor{pink!25}mIoU & \cellcolor{cyan!10}RMSE \\
        \midrule 
        \multirow{4}{*}{\rotatebox[origin=c]{90}{w/o UDA}} 
         & STL \textit{target} & DeepLabV2 & 67.93 & 6.62 & 77.10 & 6.62 \\
         & STL \textit{source} & DeepLabV2 & 35.63 & 13.79 & 58.77 & 11.99 \\
         & MTL \textit{target} & DeepLabV2 & 70.43 & 6.79 & 79.63 & 6.72 \\
         & MTL \textit{source} & DeepLabV2 & 15.32 & 14.51 & 49.50 & 12.26 \\
        \midrule
        \multirow{8}{*}{\rotatebox[origin=c]{90}{w/ UDA}} 
         & STL-UDA~\cite{vandenhende2021multi} & DeepLabV2 & 37.55 & 14.26 & 61.60 & 11.45 \\
         & MTL-UDA~\cite{vandenhende2021multi} & DeepLabV2 & 16.71 & 14.47 & 57.26 & 11.85 \\
         & XTAM~\cite{lopes2023cross} & DeepLabV2 & 37.93 & 11.66 & 63.76 & 11.15 \\
         & Swin-UDA~\cite{liu2021swin} & Swin & 39.00 & 11.03 & 63.88 & 11.09 \\
         & MulT~\cite{bhattacharjee2022mult} & Swin & 42.12 & 9.55 & 66.12 & 10.35 \\
         & VTAGML~\cite{bhattacharjee2023vision} & Swin & 50.03 & 6.99 & \textbf{70.93} & 8.66 \\
         \cmidrule(r){2-7}
         & FAMDA (Ours) & DeepLabV2 & \underline{57.02} & \textbf{6.29} & 65.86 & \textbf{6.35} \\
         & FAMDA (Ours) & SegFormer & \textbf{60.00} & \underline{6.49} & \underline{68.40} & \underline{6.46} \\
        \bottomrule
    \end{tabular}}
    \vspace{-0.4cm}
\end{table}

\begin{table*}[t]
\centering
\renewcommand{\arraystretch}{1.0}
\fontsize{7.2pt}{7.45pt}\selectfont
\caption{
\textbf{Ablations on synthetic-to-real multi-task UDA benchmarks.}
(a) \update{Controlled ablation of foundation-model guidance on VK2$\rightarrow$CS.
Here, No VFM denotes w/o SAM and w/o DAM; DAM-only denotes w/o SAM and w/ DAM; FAMDA uses both SAM and DAM.}
(b) Single-task versus multi-task training on VK2$\rightarrow$CS with MiT-B5.
(c) Three-task segmentation-depth-normal (S-D-N) performance on SYN$\rightarrow$CS.
Metrics are mIoU (\%, $\uparrow$), RMSE (m, $\downarrow$), and surface-normal mean angular error mErr. ($^\circ$, $\downarrow$).
}
\vspace{-0.15cm}
\label{tab:benchmark_ablations}

\setlength{\fboxsep}{3pt}
\setlength{\fboxrule}{0.5pt}
\def\ablationboxheight{2.4cm}

\fcolorbox{black!35}{black!2}{%
\begin{minipage}[t][\ablationboxheight][t]{0.38\textwidth}
\vspace{0pt}
\centering
\textbf{(a) FM guidance on VK2$\rightarrow$CS}\\[0.05cm]
\setlength{\tabcolsep}{2.5pt}
\begin{tabular}{@{}l|cc|cc|cc@{}}
\toprule
\multirow{2}{*}{Backbone}
& \multicolumn{2}{c|}{No VFM}
& \multicolumn{2}{c|}{DAM-only}
& \multicolumn{2}{c}{FAMDA} \\
\cmidrule(r){2-3} \cmidrule(r){4-5} \cmidrule(l){6-7}
& \cellcolor{pink!25}mIoU
& \cellcolor{cyan!10}RMSE
& \cellcolor{pink!25}mIoU
& \cellcolor{cyan!10}RMSE
& \cellcolor{pink!25}mIoU
& \cellcolor{cyan!10}RMSE \\
\midrule
MiT-B0 & 49.83 & 11.08 & 50.75 & 6.73 & 60.97 & 6.77 \\
MiT-B1 & 51.61 & 10.35 & 47.90 & 6.51 & 67.00 & 6.52 \\
MiT-B2 & 49.08 &  9.79 & 56.69 & 6.46 & 65.05 & \textbf{6.34} \\
MiT-B5 & 49.69 & 10.82 & 57.45 & \textbf{6.34} & \textbf{68.40} & 6.46 \\
\bottomrule
\end{tabular}
\end{minipage}}
\hfill
\fcolorbox{black!35}{black!2}{%
\begin{minipage}[t][\ablationboxheight][t]{0.265\textwidth}
\vspace{2pt}
\centering
\textbf{(b) Single-task vs. multi-task}\\[0.15cm]
\setlength{\tabcolsep}{2.5pt}
\begin{tabular}{@{}l|cc|cc@{}}
\toprule
\multirow{2}{*}{Model}
& \multicolumn{2}{c|}{FM}
& \multicolumn{2}{c}{VK2$\rightarrow$CS} \\
\cmidrule(r){2-3} \cmidrule(l){4-5}
& SAM & DAM
& \cellcolor{pink!25}mIoU
& \cellcolor{cyan!10}RMSE \\
\midrule
Seg. only   & \cmark & \xmark & 64.48 & -- \\
Depth only & \xmark & \cmark & --    & \textbf{6.38} \\
Multi-task & \cmark & \cmark & \textbf{68.40} & 6.46 \\
\bottomrule
\end{tabular}
\end{minipage}}
\hfill
\fcolorbox{black!35}{black!2}{%
\begin{minipage}[t][\ablationboxheight][t]{0.28\textwidth}
\vspace{2pt}
\centering
\textbf{(c) Three-task S-D-N on SYN$\rightarrow$CS}\\[0.15cm]
\setlength{\tabcolsep}{3.0pt}
\begin{tabular}{@{}l|ccc@{}}
\toprule
Backbone
& \cellcolor{pink!25}mIoU
& \cellcolor{cyan!10}RMSE
& \cellcolor{green!15}mErr. \\
\midrule
MiT-B0    & 51.34 & 6.97 & 38.21 \\
MiT-B1    & 56.28 & 6.79 & 38.59 \\
MiT-B2    & 59.66 & 6.44 & \textbf{38.04} \\
MiT-B5    & \textbf{60.62} & 6.48 & 38.05 \\
DeepLabV2 & 57.80 & \textbf{6.38} & 38.21 \\
\bottomrule
\end{tabular}
\end{minipage}}

\vspace{-0.3cm}
\end{table*}

\subsubsection{Computational Efficiency and Embedded Runtime}
\label{sec:runtime}

Efficient inference is important for fieldable multi-task perception because separate segmentation and depth models increase model size, runtime, and power consumption. Tab.~\ref{tab:lowlight} shows the model-level accuracy-efficiency tradeoff against foundation-model and single-task baselines. Direct foundation-model inference is impractical for the target setting: SSAM-H has very high latency, while DAM-L provides only depth. Running separate single-task UDA models also increases cost because two networks are required to obtain both outputs. In contrast, FAMDA produces semantic segmentation and depth in one forward pass.


To evaluate deployment feasibility, we export the trained student to ONNX, convert it to TensorRT FP16 on the Jetson Orin Nano 8GB, and report runtime across backbones in Tab.~\ref{tab:embedded_runtime}. The similar mIoU/RMSE values in Tabs.~\ref{tab:lowlight} and~\ref{tab:embedded_runtime} indicate that TensorRT deployment preserves low-light accuracy while changing runtime, memory, and power profile.


Tab.~\ref{tab:embedded_runtime} shows that all FAMDA variants operate within a low-power embedded regime on the Jetson Orin Nano. Ours-B0 achieves the highest throughput at 8.1 FPS and 7.28 W board power, while Ours-B5 improves accuracy at the cost of lower throughput and higher memory use. Ours-B2 offers a strong deployment point, achieving 55.01 mIoU and 5.53 m RMSE at 6.3 FPS with 7.75 W board power. Compared with running separate segmentation and depth models, the shared FAMDA student provides both outputs in a single embedded inference pass, which is important for fieldable low-light perception under size, weight, power, and latency constraints. \update{A representative Ours-B2 component-level timing breakdown, including raw-frame preprocessing and simple display-resolution colorization, is provided in the supplementary material (Tab.~S12).}


\subsection{Standard Synthetic-to-Real Benchmarks}
\label{sec:standard_benchmarks}

Following standard multi-task UDA protocols~\cite{lopes2023cross,bhattacharjee2023vision}, we evaluate on SYNTHIA$\rightarrow$Cityscapes (SYN$\rightarrow$CS) and Virtual KITTI2$\rightarrow$Cityscapes (VK2$\rightarrow$CS). These benchmarks are used for comparison with prior multi-task UDA methods because established evaluations with both semantic segmentation and depth are primarily synthetic-to-real.

Tab.~\ref{tab:main_comparison} shows that FAMDA achieves state-of-the-art or competitive performance across multi-task UDA benchmarks. On SYN$\rightarrow$CS, FAMDA obtains the best segmentation mIoU and strong depth performance compared to prior baselines. On VK2$\rightarrow$CS, FAMDA achieves the best depth RMSE and competitive segmentation mIoU. These results indicate that the proposed foundation-model-assisted self-training strategy is competitive with specialized multi-task UDA architectures while providing a deployment-oriented adaptation pipeline. The fully supervised target-domain rows provide upper-bound references, while the source-only rows quantify the domain gap without adaptation.


\subsection{Ablation Studies}
\label{sec:ablations}
Tab.~\ref{tab:benchmark_ablations} summarizes three ablation studies. \update{First, Tab.~\ref{tab:benchmark_ablations}(a) isolates foundation-model guidance on VK2$\rightarrow$CS. The no-VFM baseline disables both SAM refinement and DAM pseudo-depth, using DAFormer-style semantic self-training with source-domain depth labels for the depth head. Across B0--B5, adding DAM reduces RMSE from 9.79--11.08 m to 6.34--6.73 m, while adding SAM on top mainly improves semantic adaptation, raising mIoU by +8.36 to +19.10.}

Second, the single-task comparison in Tab.~\ref{tab:benchmark_ablations}(b) shows that FAMDA's multi-task student provides both outputs from one model while remaining competitive with separate task-specific models. On VK2$\rightarrow$CS, the multi-task MiT-B5 model achieves higher segmentation accuracy than the single-task segmentation model and depth RMSE close to the single-task depth model.

Finally, Tab.~\ref{tab:benchmark_ablations}(c) evaluates a three-task extension with semantic segmentation, depth estimation, and surface normal prediction on SYN$\rightarrow$CS. FAMDA is extended by attaching an additional normal decoder head without modifying the core UDA framework. The three-task models preserve strong segmentation and depth performance while producing competitive surface-normal estimates, suggesting that the proposed foundation-model-assisted self-training strategy can scale beyond the two-task setting. 
\section{Conclusion}

We presented FAMDA, a framework that integrates vision foundation models into self-training-based UDA for multi-task dense prediction. By refining segmentation pseudo-labels with SAM and using DAM for depth supervision, FAMDA distills foundation-model guidance into a lightweight student network. Across standard synthetic-to-real multi-task UDA benchmarks and our field-collected daytime-to-low-light dataset, FAMDA achieves strong segmentation and depth accuracy with favorable accuracy-efficiency tradeoffs. We further instantiate the approach on a handheld DomiNite-based low-light prototype and demonstrate TensorRT inference on Jetson Orin Nano, supporting practical onboard multi-task perception under low-power constraints. These results suggest a promising direction for fieldable low-light perception systems requiring both semantic and geometric scene understanding. 

\vspace{0.1cm}
\noindent\textbf{Acknowledgments.}
We would like to thank Kevin Kaighn, Colin Earle, Jesse Battaglia, Josh Chattin, and Abhinav Rajvanshi for helping with the low-light system prototype and data collection.

{
    \small
    \bibliographystyle{ieeenat_fullname}
    \bibliography{main}
}

\end{document}